\documentclass[a4paper,fleqn]{cas-dc}

\usepackage[numbers]{natbib}

\def\tsc#1{\csdef{#1}{\textsc{\lowercase{#1}}\xspace}}
\tsc{WGM}
\tsc{QE}
\tsc{EP}
\tsc{PMS}
\tsc{BEC}
\tsc{DE}

\begin{document}
%
%
%
\let\printorcid\relax
\title [mode = title]{Ensemble Predicate Decoding for Unbiased Scene Graph Generation}                      
\author[1,2]{\textcolor{black}{Jiasong Feng}}
\author[1,2]{\textcolor{black}{Lichun Wang}}
\cormark[1]
\author[1,2]{\textcolor{black}{Hongbo Xu}}
\author[1,2]{\textcolor{black}{Kai Xu}}
\author[1,2]{\textcolor{black}{Baocai Yin}}

\affiliation[1]{organization={Beijing Key Laboratory of Multimedia and Intelligent Software Technology},
	addressline={Beijing Artificial Intelligence Institute}, 
	city={Beijing},
	postcode={100124}, 
	country={China}}
\affiliation[2]{organization={School of Information Science and Technology},
	addressline={Beijing University of Technology}, 
	city={Beijing},
	postcode={100124}, 
	country={China}}
	
\cortext[cor1]{Corresponding author}
\nonumnote{Email address: wanglc@bjut.edu.cn (Lichun Wang)}
\begin{abstract}
	Scene Graph Generation (SGG) aims to generate a comprehensive graphical representation that accurately captures the semantic information of a given scenario. However, the SGG model's performance in predicting more fine-grained predicates is hindered by a significant predicate bias. According to existing works, the long-tail distribution of predicates in training data results in the biased scene graph. However, the semantic overlap between predicate categories makes predicate prediction difficult, and there is a significant difference in the sample size of semantically similar predicates, making the predicate prediction more difficult. Therefore, higher requirements are placed on the discriminative ability of the model. In order to address this problem, this paper proposes Ensemble Predicate Decoding (EPD), which employs multiple decoders to attain unbiased scene graph generation. Two auxiliary decoders trained on lower-frequency predicates are used to improve the discriminative ability of the model. Extensive experiments are conducted on the VG, and the experiment results show that EPD enhances the model's representation capability for predicates. In addition, we find that our approach ensures a relatively superior predictive capability for more frequent predicates compared to previous unbiased SGG methods.
\end{abstract}

\maketitle
\section{Introduction}
The primary objective of scene graph generation (SGG) is to construct a structured representation that effectively captures the semantic information of scenario \cite{johnson2015image}. Specifically, objects are depicted as nodes, while visual relationships between objects are established through edges within the scene graph. SGG has been proven to be valuable in various vision downstream tasks, including image captioning \cite{song2021spatial,chen2020say}, visual question answering \cite{tang2019learning,lu2016hierarchical}, video understanding\cite{nguyen2024hig,han2022one}, and image retrieval \cite{zeng2021conceptual}.

\begin{figure}[!h]
	\centering
	\includegraphics[width=0.89\linewidth]{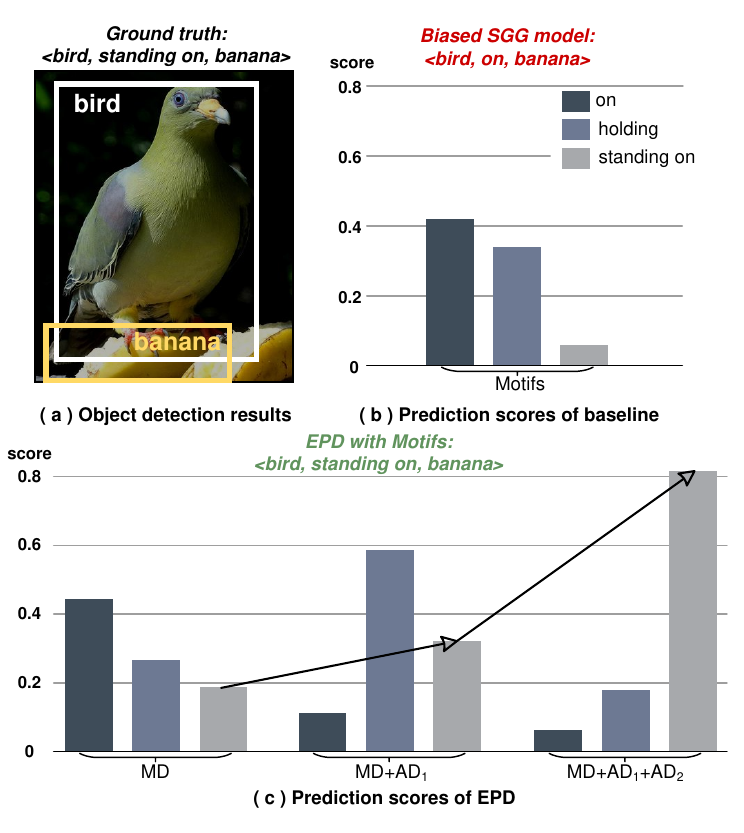}
	\caption{Scores of top-3 predictions for the triplet $\langle \textit{bird}, \textit{--}, \textit{banana} \rangle$ across different SGG models, which ground truth is \textit{standing on}. Motifs is affected by bias, and the coarse predicate \textit{on} receives the highest score.  The proposed Ensemble Predicate Decoding (EPD) predicts the triplet as $\langle \textit{bird}, \textit{standing on}, \textit{banana} \rangle$. EPD includes three decoders, the main decoder $\mathcal{MD}$, the auxiliary decoders $\mathcal{AD}_1$ and $\mathcal{AD}_2$.}
	\label{figure1}
	\vspace{-0.8cm}
\end{figure}

Although significant progress has been made in the field of scene graph generation in recent years \cite{lin2020gps,yang2021probabilistic}, SGG still suffers from bias\cite{tang2020unbiased}. Specifically, the bias leads to the model's tendency to predict coarse-grained predicates, resulting in fragmented scene graph information that cannot be effectively applied to downstream tasks. Fig. \ref{figure1} shows the predicted results of different SGG models. In Fig. \ref{figure1} (b), due to the influence of bias, the biased SGG model Motifs \cite{dong2022stacked, li2022ppdl,herzig2020learning} incorrectly predicts the relationship of the triplet $\langle \textit{bird}, \textit{--}, \textit{banana} \rangle$ with a head category predicate \textit{on}, instead of the ground truth predicate \textit{standing on}. Previous works  \cite{dong2022stacked, li2022ppdl,herzig2020learning} have attributed this to the long-tailed distribution of predicates in the dataset, and substantial efforts have been dedicated to addressing the issue. However, GCL \cite{dong2022stacked} found that for a naive SGG model, whether conventional or debiased, it could only distinguish a limited range of predicates with relatively equal amounts of training instances. They attribute this limitation to the constrained discriminative capacity of single classifiers on predicates representations. Based on this discovery, GCL developed a novel group collaborative learning strategy, utilizing multiple classifiers to improve the classification performance of the model.

In the dataset used to train SGG models, another phenomenon that coexists with predicate imbalance is that the semantics of two predicates with significant differences in sample size are similar. For instance, predicates \textit{standing on} and \textit{on} can be defined as a semantically proximate pair, because their representations learned by models exhibit significant similarity \cite{chen2023addressing}. Meanwhile, during the SGG training phase, \textit{standing on} appears 8,768 times and \textit{on} appears 429,891 times. The multiple classifier scheme can effectively expand the discriminative capacity of the model, but it does not provide a targeted solution for discriminating semantically similar predicates. The compounded effect of similarity and long-tailed factors poses a catastrophic challenge for predicate predictions.

To cope with the challenge, we propose employing multiple decoders to enhance the model's ability to decode predicate representations. In detail, we train one main decoder $\mathcal{MD}$ and two auxiliary decoders $\mathcal{AD}_1$ and $\mathcal{AD}_2$. The three decoders are trained on distinct subsets of the training set. one on the entire dataset, one focusing on body-tail predicates, and another on tail predicates exclusively. $\mathcal{MD}$ is trained on the subset including all predicates. $\mathcal{AD}_1$ is trained on the subset that includes body predicates and tail predicates. $\mathcal{AD}_2$ is trained on the subset that only includes the tail predicates. Compared to high-frequency predicates, low-frequency predicates are used more by decoders in the training process, which helps to reduce the impact of long-tail effects on predicate prediction. In addition, predicate pairs with similar semantics, such as \textit{standing on} and \textit{on}, are separated during specific decoder training processes. For decoders trained on the subset that only contains tail predicates, the predicate \textit{on} is invisible, which can reduce the confusion of the model in understanding similar predicates. Fig. \ref{figure1}(c) shows that, by combining more auxiliary decoders with the main decoder, the prediction of the triplet $\langle \textit{bird}, \textit{--}, \textit{banana} \rangle$ is gradually refined, from $\langle \textit{bird}, \textit{on}, \textit{banana} \rangle$ to $\langle \textit{bird}, \textit{holding}, \textit{banana} \rangle$, and then further refined to $\langle \textit{bird}, \textit{standing on}, \textit{banana} \rangle$.This paper makes two contributions:
\begin{itemize}
	\item Propose a new model-agnostic SGG approach EPD (Ensemble Predicate Decoding), which trains one main decoder and two auxiliary decoders. The main decoder is responsible for decoding all predicates. The auxiliary decoders focus on enhancing the model’s decoding capacity for specific infrequent predicates.
	\item Extensive experiments are performed on the widely used SGG benchmark dataset, Visual Genome (VG). EPD achieves exceptional performance when combined with various scene graph baselines. Specifically, when using EPD, mR@K has significantly improvement and R@K only shows a minimal decrease. Experimental results show that our method outperforms the state-of-the-art approach.
\end{itemize}

\section{Related Work}
\label{sec:formatting}
\noindent\textbf{Unbiased Scene Graph Generation.}
SGG has made significant progress in recent years \cite{li2024scene}. However, due to the bias in the distribution of predicates in the dataset, the practical application of these models is still limited\cite{tang2020unbiased,yu2020cogtree}. In order to tackle the issues arising from the long-tail effect of predicates, researchers have introduced various unbiased SGG models\cite{li2022ppdl,tong2023alleviating,li2023label,peng2024causality}. PPDL\cite{li2022ppdl} suggested an improved re-weighting strategy based on the similarity between the predicted probability distribution and the ground truth distribution. LS-KD \cite{li2023label} alleviated the training bias by dynamically generating a soft label for each subject-object instance through the fusion of predicted label semantic distribution with its original one-hot target label. DDloss \cite{peng2024causality} reduces bias by leveraging a causal graph approach, improving the fairness in relation predication. The existing SGG model overlooked the relationship between the model's classification ability and its classification range. Some works are dedicated to improving the discriminative ability of classifiers. MED\cite{tong2023alleviating} proposed a multi-expert predicate module to weaken the label ambiguity problem in SGG, which includes two models that learn from the full label set and the implicit label set respectively. GCL \cite{dong2022stacked} introduced a group cooperative learning strategy to train multiple classifiers on a set of relatively balanced predicate groups. However, although the multiple classifier scheme can effectively expand the discriminative capacity of the model, it lacks consideration for discriminating semantically similar predicates.

\noindent\textbf{Ensemble Learning.}
The idea of ensemble learning is to employ multiple base learners and combine their prediction\cite{polikar2012ensemble,geiselhart2021automorphism}. Choosing appropriate training data configurations for learners in different contexts is crucial. One common configuration is that base learners take the entire training set as input, as demonstrated in RAMED\cite{shen2021time}. When using this kind of configuration, base learners generate outputs with varying dimensions according to a prescribed pattern (e.g., from coarse-grained to fine-grained), which are subsequently harmonized through specific strategies. Another configuration is that each base learner takes input from one subset of the complete training set, such as bagging strategy\cite{garcia2009constructing}. Training on a specially selected subset allows the base learners to focus more on specified learning tasks.

For SGG, due to the effects of semantic overlap and the long-tail distribution, low-frequency predicates are more difficult to decode\cite{goel2022not}. Therefore, following the idea of using multiple classifiers\cite{zhang2022self,cui2022reslt}, we set up multiple decoders and provide the second configuration mentioned above for decoder training, ensuring that different predicate decoders concentrate on issues of different difficulty levels.

\section{Methodology}
\subsection{Problem Definition}
Given an input image, the objective of SGG is to transform it into the form of a scene graph $\mathcal{G}=\left ( \mathcal{N},\mathcal{E} \right )$, where each element of the node set $\mathcal{N}$ corresponds to an object in the image, and each element of the edge set $\mathcal{E}$ corresponds to the relation between a pair of objects. The relationship between object $n_i\in\mathcal{N}$ and $n_j\in\mathcal{N}$ is denoted as $e_{ij} \in \mathcal{E}$, then the triplet $\langle \textit{$n_{i}$}, \textit{$e_{ij}$}, \textit{$n_{j}$} \rangle$ , is
an instance of the triplet $\langle \textit{subject}, \textit{predicate}, \textit{object} \rangle$.

\subsection{Overview}
\begin{figure*}
	\centering
	\includegraphics[scale=0.55]{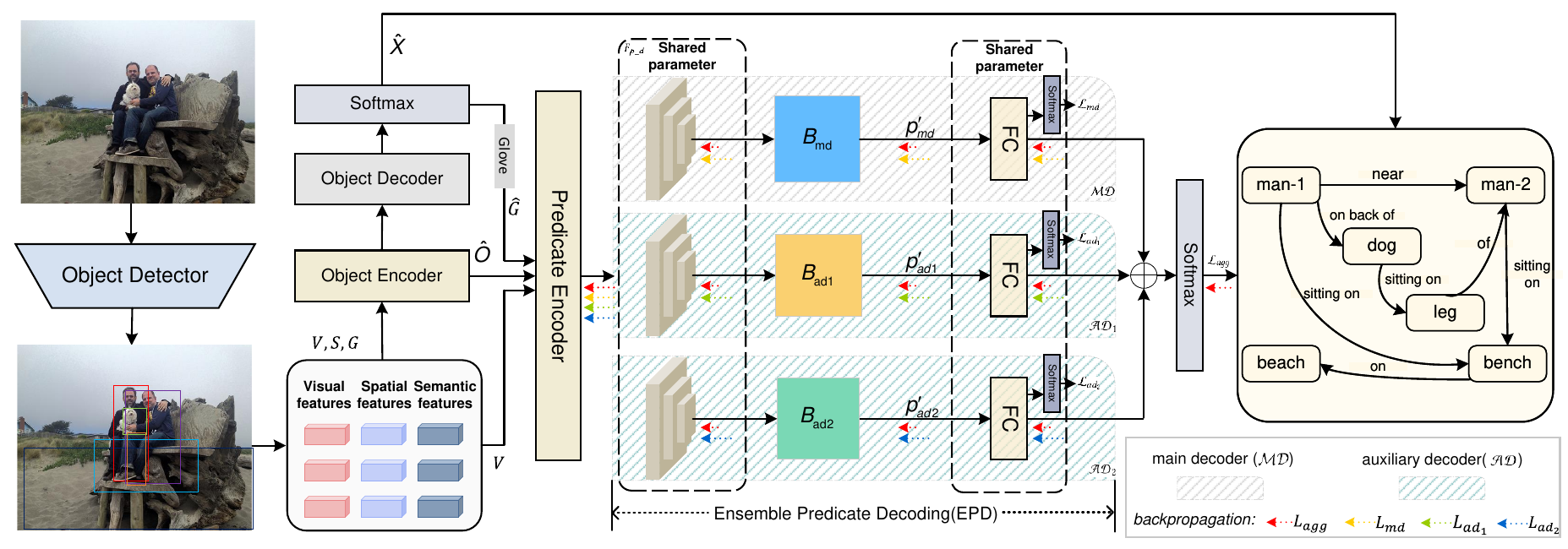}
	\caption{Overview of SGG model using EPD. In the ensemble predicate decoding stage, we process predicate encoding features with multiple decoders. One decoder acts as the main, and the others as auxiliary decoders. Each decoder branch shares parameters partially, generating predicate decoding features $p'_{md}$, $p'_{ad_1}$, $p'_{ad_2}$. After passing through a shared predicate classifier, the decoders' outputs are predicted and subsequently integrated to obtain the result. }
	\label{figure2}
\end{figure*}

Fig. \ref{figure2} shows the framework of the proposed EPD. Image features are first extracted by utilizing a backbone network, such as ResXNet-101 \cite{he2016deep}, and then processed by an object detector like Faster R-CNN \cite{ren2015faster} to get proposals. For each proposal, the detector predicts bounding box coordinates, object labels, and visual features. The coordinates are embedded into spatial features, and the labels are embedded in semantic features. These features, along with the visual features, are concatenated and passed through an object encoder to generate refined object features $\hat{O}=\left \{\hat{o}_{1},...,\hat{o}_{n} \right \}$. The refined object features are then fed into the object decoder to obtain refined labels $\hat{X}=\left \{\hat{x}_{1},...,\hat{x}_{n} \right \}$, which are labels of the nodes in the scene graph.

Meanwhile, the refined object labels $\hat{X}$ are embedded into refined semantic features. Then, the refined object features, the refined semantic features, and visual features are input into the predicate encoder, generating predicate encoding features corresponding to the objects. The predicate encoding features of each pair of objects, combined with the visual features extracted from the union of their bounding boxes, form the predicate features representing their relationship. These predicate features are then fed into the main decoder $\mathcal{MD}$ and the auxiliary decoders $\mathcal{AD}_1$ and $\mathcal{AD}_2$. Each decoder branch has its own parameters, generating predicate decoding features $p'_{md}$, $p'_{ad_1}$, $p'_{ad_2}$, which are then fed into the predicate classifier to obtain three decoder logits. The logits from decoders are then ensembled and used to predict predicate labels. Finally, the scene graph $\mathcal{G}$ is constructed by using the refined object labels as nodes, while the predicate labels are employed as edges.

\subsection{Object Encoding and Decoding}
For a single image with $n$ objects, the object detector predicts the bounding boxes $B=\{b_1,...,b_n\}$, visual features $V=\{v_1,...,v_n\}$, and object labels $X=\{x_1,...,x_n\}$. Then, the elements of set $B$ are passed through the dimensionality-increasing embedding layer $W_b(\cdot)$ to obtain spatial features $S=\{s_1,...,s_n\}$. And the elements of set $X$ are embedded by Glove embedding layer to obtain semantic features $G=\left \{g _{1},...,g_{n} \right \}$. Taking the $i^{th}$ object as an example, the object encoder leverages the visual, spatial, and semantic features to obtain refined object features $\hat{o}_i$, which are formulated as:
\begin{equation}
	\hat{o}_i = F_{o\_e}([v_i;s_i;g_i])
	\label{eq:obj_encoder}
\end{equation}
where $v_i\in V$, $s_i\in S$, and $g_i\in G$ are the visual, spatial, and semantic features of the $i^{th}$ object. $F_{o\_e}(\cdot)$ denotes the linear transformation function of the object encoder. The notation $[ \cdot ; \cdot ] $ represents the concatenation operation.

During the object decoding stage, the object decoder takes $\hat{O}=\{ \hat{o}_1,...,\hat{o}_n\}$ as input and outputs refined object labels $\hat{X}=\{ \hat{x}_1,...,\hat{x}_n\}$. The elements of set $\hat{X}$ are also embedded by Glove embedding layer to obtain refined semantic features $\hat{G}=\left \{\hat{g}_1,...,\hat{g}_n \right \}$, which will be used for predicate encoding.

\subsection{Predicate Encoding}
In the predicate encoding stage, the SGG model concatenates refined object feature $\hat{o}_i$ with the visual feature $v_i$ and refined semantic feature $\hat{g}_i$. The concatenated feature serves as raw predicate representation for the $i^{th}$ object and is fed into the predicate encoder to obtain the predicate encoding feature $p_i$, as described in the following:
\begin{equation}
	p_i = F_{p\_e}([\hat{g}_i;\hat{o}_i;v_i])
	\label{eq:rel_encoder}
\end{equation}
$F_{p\_e}(\cdot)$ refers to the linear transformation function of the predicate encoder.
\subsection{Ensemble Predicate Decoding}
As illustrated in Fig. \ref{figure2}, ensemble predicate decoding is achieved by integrating three decoders: main decoder $\mathcal{MD}$ and two auxiliary decoders $\mathcal{AD}_{1}$, $\mathcal{AD}_{2}$. The three decoders are trained on different training subsets. In the following description, it is assumed that the relationship between the $i^{th}$ object and the $j^{th}$ object is represented by the $k^{th}$ predicate.

First, the model extracts the visual features of the merged area of the bounding boxes of two objects $o_i$ and $o_j$, denoted as union feature $u_{ij}$. Then, the predicate encoding features $p_i$, $p_j$ and the union feature $u_{ij}$ are fed to decoders to obtain predicate decoding features as follows:
\begin{equation}
	\begin{split}
		p'_{md} &= B_{md}(F_{p\_d}(W_p([{p_i};{p_j}])*u_{ij}))
		\\
		p'_{ad_1} &= B_{ad_1}(F_{p\_d}(W_p([{p_i};{p_j}])*u_{ij}))
		\\
		p'_{ad_2} &= B_{ad_2}(F_{p\_d}(W_p([{p_i};{p_j}])*u_{ij}))
		\\
	\end{split}
\end{equation}
where $p'_{md}$, $p'_{ad_1}$ and $p'_{ad_2}$ are the predicate decoding features output by the main decoder and the two predicate auxiliary decoders, respectively. $W_p(\cdot)$ is a linear layer used for dimension expansion. $F_{p\_d}(\cdot)$ denotes the shared linear transformation function. $B_{md}(\cdot)$, $B_{ad_1}(\cdot)$, and $B_{ad_2}(\cdot)$ are batch normalization layers that are independent, each with its own learnable parameters.

Finally, the predicate decoding features $p'_{md}$, $p'_{ad_1}$ and $p'_{ad_2}$ are input to a fully connected layer to obtain three groups of predicted logits of the predicate:
\begin{equation}
	\begin{aligned}
		z_{md} &= FC(p'_{md})\\
		z_{ad_1} &= FC(p'_{ad_1})\\
		z_{ad_2} &= FC(p'_{ad_2})\\
	\end{aligned}
\end{equation}

In order to make the model give more attention to low-frequency predicates, all three decoders are involved in learning low-frequency predicates. Specifically, for low-frequency predicates, losses are calculated based on the predictions obtained from three decoders and used for backpropagation. To improve the decoding ability of the model, the three decoders are required to focus on learning predicates within different ranges, which is achieved by training the three decoders with different sample sets. Specifically, the training samples are divided into three parts based on predicate frequency: head predicate, body predicate, and tail predicate, as outlined in previous work \cite{li2022ppdl}.

Denote the set containing all head predicate samples as $N_1$, the set containing all body predicate samples as $N_2$, and the set containing all tail predicate samples as $N_3$. The union of $N_1$, $N_2$, and $N_3$ is the complete set of training samples, that is, $N={N_1}\cup{N_2}\cup{N_3}$. The complete set of training samples $N$ is assigned to the main decoder $\mathcal{MD}$, denoted as $N_{md}$. The training sample subset ${N_2}\cup{N_3}$ is assigned to auxiliary decoder $\mathcal{AD}_1$, denoted as $N_{ad_1}$. The training sample subset $N_3$ is assigned to auxiliary decoder $\mathcal{AD}_2$, denoted as $N_{ad_2}$. The loss functions for decoders are described as follows:
\begin{equation}
	\begin{split}
		&\mathcal{L}_{md} = \frac{1}{|N_{md}|} \sum_{k\in N_{md}}CE(sc_{md}^{k},\hat{y}_{k} )\\
		&\mathcal{L}_{ad_1} = \frac{1}{|N_{ad_1}|} \sum_{k\in N_{ad_1}} CE(sc_{ad_1}^{k},\hat{y}_{k} )\\
		&\mathcal{L}_{ad_2} = \frac{1}{|N_{ad_2}|} \sum_{k\in N_{ad_2}} CE(sc_{ad_2}^{k},\hat{y}_{k} )
	\end{split}
\end{equation}
where $sc_{md}^{k}$, $sc_{ad_1}^{k}$, and $sc_{ad_2}^{k}$ are obtained by applying the softmax function to $z_{md}^{k}$, $z_{ad_1}^{k}$, and $z_{ad_2}^{k}$, $\hat{y}_k$ is the ground truth predicate label in the form of one-hot vector.

In order to ensure collaborative prediction among the three decoders and consistency settings between the training and inference phases, we also employ an aggregated loss $\mathcal{L}_{agg}$, which is calculated based on the aggregated prediction score$sc_{sum}^{k}$:
\begin{equation}
	\begin{split}
		\mathcal{L}_{agg} = \frac{1}{|N|} \sum_{k\in N} CE(sc_{sum}^{k},\hat{y}_{k} )
	\end{split}
\end{equation}
where $sc_{sum}^{k}$ is obtained by applying the softmax function to the aggregated prediction $z_{sum}$ of the three decoders. $z_{sum}$ is computed as follows:
\begin{equation}
	\begin{split}
		&z_{sum} = \lambda_{md}z_{md}+\lambda_{ad_1}z_{ad_1} +\lambda_{ad_2}z_{ad_2}
		\label{eq7}
	\end{split}
\end{equation}
where $\lambda_{md}$, $\lambda_{ad_1}$ and $\lambda_{ad_2}$ are weighting factors corresponding to $\mathcal{MD}$, $\mathcal{AD}_1$ and $\mathcal{AD}_2$ .
Due to training on the complete dataset, the main decoder is more robust. Therefore, it is necessary to ensure that it is in a dominant position and receives effective support from the auxiliary decoders. So, among the weights of the three decoders, we specify that the weight of the main decoder is the maximum value.  For the weights of two auxiliary decoders, considering that the subset used for training $\mathcal{AD}_2$ includes the most biased samples of the training set, $\lambda_{ad_2}$ is assigned a larger value than $\lambda_{ad_1}$. As a result, the assignment  $\lambda_{md} \textgreater \lambda_{ad_2} \textgreater \lambda_{ad_1}$  aims to maintain a balance between the three decoders.
During the inference phase, the aggregated prediction is the final prediction.

The total loss is the combination of the aggregated loss and each decoder's individual loss :
\begin{equation}
	\begin{split}
		\mathcal{L}_{all} = (1-\gamma)( \mathcal{L}_{md} + \alpha \mathcal{L}_{ad_1} +  \beta \mathcal{L}_{ad_2}) + \gamma \mathcal{L}_{agg}
		\label{eq8}
	\end{split}
\end{equation}
where the hyperparameters $\alpha$ and $\beta$ are used to adjust attention to errors from auxiliary decoders during training, and $\gamma$ is utilized to balance the decoding ability of individual decoders and the collaborative ability of the three decoders.

\section{Experiments}
In this section, we first assess the performance of the proposed EPD on the widely-used VG dataset for scene graph generation. Then, ablation experiments and hyperparameter experiments are conducted. The final visualization experiment demonstrates that EPD does not excessively suppress the head and body predicates, which also explains why EPD achieves good Mean performance while improving the prediction accuracy of tail predicates.
\begin{table*}[!htbp]
	\begin{center}
		\caption{Performance comparison on VG. The best results are in bold. $^{\dagger}$ denotes the results are reproduced based on the code~\cite{tang2020unbiased}.}
		\resizebox{1\textwidth}{!}{
			\begin{tabular}{cccccccccc}
				\hline \hline
				\multirow{2}{*}{Methods} & \multicolumn{3}{c}{PredCls}                                                            & \multicolumn{3}{c}{SGCls}                                                              & \multicolumn{3}{c}{SGDet}                                                                                        \\
				& R@50/100  & mR@50/100                              & Mean                              & R@50/100  & mR@50/100                              & Mean                              & R@50/100  & mR@50/100                                                        & Mean                              \\ \hline \hline
				Motifs$^{\dagger}$\cite{zellers2018neural}              & 65.3/67.2 & 14.9/16.3                              & 40.9                              & 38.9/39.8 & 8.3/8.8                                & 23.9                              & \textbf{32.1}/\textbf{36.8} & 6.6/7.9                                                          & 20.8                              \\
				VCTree$^{\dagger}$\cite{tang2019learning}                   & \textbf{65.9}/\textbf{67.5} & 17.2/18.5                              & 42.2                              & \textbf{45.3}/\textbf{46.2} & 10.6/11.3                              & 28.3                              & 31.9/36.2 & 7.1/8.3                                                          & 20.8                              \\ \hline \hline
				BGNN\cite{li2021bipartite}                     & 59.2/61.3 & 30.4/32.9                              &  45.9 & 37.4/38.5 & 14.3/16.5                              & 26.6                             & 31.0/35.8 & 10.7/12.6                                                        &  22.5 \\
				PCPL\cite{yan2020pcpl}                     & 50.8/52.6 & 35.2/37.8                              & 44.1                              & 27.6/28.4 & 18.6/19.6                              & 23.5                              & 14.6/18.6 & 9.5/11.7                                                         & 13.6                              \\
				HetSGG\cite{yoon2023unbiased}                     & 57.1/59.4 & 32.3/34.5    &  45.8 & 37.6/38.5 & 15.8/17.7                              & 27.4                              & 30.2/34.5 & 11.5/13.5                                                        &  22.4 \\
				\hline \hline
				Motifs+PPDL\cite{li2022ppdl}              & 47.2/47.6 & 32.2/33.3                              & 40.0                              & 28.4/29.3 & 17.5/18.2                              & 23.3                              & 22.4/23.9 & 11.4/13.5                                                        & 17.8                              \\
				Motifs+TDE\cite{tang2020unbiased}               & 46.2/51.4 & 25.5/29.1                              & 38.0                              & 27.7/29.9 & 13.1/14.9                              & 21.4                              & 16.9/20.3 & 8.2/9.8                                                          & 13.8                              \\
				Motifs+LS-KD\cite{li2023label}             & 55.1/59.1 & 24.1/27.4                              & 41.4                              & 33.8/35.8 & 13.8/15.2                              & 24.6                              & 25.6/29.7 & 9.7/11.5     &                                                   19.1                              \\
				Motifs+DKBL\cite{chen2023addressing}               & 57.2/58.8 & 29.7/32.2                              & 44.4                              & 32.7/33.4 & 18.2/19.4                             & 25.8                              & 27.0/30.7 & 12.6/15.1 & 21.3                              \\
				Motifs+Inf\cite{biswas2023probabilistic}               & 51.5/55.1 & 24.6/30.7                              & 40.3                              & 32.2/33.8 & 14.5/17.4                             & 24.5                              & 23.9/27.1 & 9.4/11.7 & 18.0                              \\
				Motifs+GCL\cite{dong2022stacked}               & 42.7/44.4 & 36.1/38.2                              & 40.4                              & 26.1/27.1 & 20.8/21.8                              & 24.0                              & 18.4/22.0 &  16.8/19.3 & 19.1                              \\
				Motifs+EPD(ours)         & 54.1/56.0 &  36.3/38.8 &  \textbf{46.3}    & 30.8/31.9 & 21.2/22.4                              & 26.5                              & 29.5/31.5 &  \textbf{17.3}/\textbf{19.0} & \ \textbf{24.3}    \\ \hline
				VCTree+PPDL\cite{li2022ppdl}              & 47.6/48.0 & 33.3/33.8                              & 40.6                              & 32.1/33.0 & 21.8/22.4                              & 27.3                              & 20.1/22.9 & 11.3/13.3                                                        & 16.9                              \\
				VCTree+TDE\cite{tang2020unbiased}               & 47.2/51.6 & 25.4/28.7                              & 38.2                              & 25.4/27.9 & 12.2/14.0                              & 19.9                              & 19.4/23.2 & 9.3/11.1                                                         & 15.8                              \\
				VCTree+LS-KD\cite{li2023label}             & 55.7/59.5 & 24.2/27.1                              & 41.6                              & 38.4/40.5 & 17.3/19.1                              & 28.8 & 25.3/29.5 & 9.7/11.3                                                         & 19.0                              \\
				VCTree+DKBL\cite{chen2023addressing}               & 60.1/61.8 & 28.7/31.3    & 45.4                              & 38.8/39.7 & 21.2/22.6    & 30.5  & 26.9/30.7 & 11.8/14.2 &20.9     \\
				VCTree+Inf\cite{biswas2023probabilistic}               & 59.5/61.0 & 28.1/30.7    & 44.8                              & 40.7/ 41.6 & 17.3/19.4    & \textbf{31.7}  & 27.7/30.1 & 10.4/11.9 &20.0     \\
				VCTree+GCL\cite{dong2022stacked}               & 40.7/42.7 & \textbf{37.1}/\textbf{39.1}    & 39.9                              & 27.7/28.7 & 22.5/23.5                              & 25.6                              & 17.4/20.7 & 15.2/17.5                                                        & 17.7                              \\
				VCTree+EPD(ours)         & 55.4/58.7 & 32.0/35.2                              & 45.3                              & 33.1/35.4 & \textbf{22.9}/\textbf{25.0 }& 29.1    & 21.7/25.4 & 13.8/16.3                                                        & 19.3                              \\ \hline \hline
			\end{tabular}
			\label{tab1}
		}
	\end{center}
	
\end{table*}
\subsection{Dataset and Evaluation Criteria}

\noindent\textbf{Dataset} We evaluate our approach on commonly used
large-scale VG benchmark \cite{krishna2017visual}, which comprises 108,077 images annotated with 75,000 object categories and 40,000 predicate categories. Following the prior studies \cite{li2022ppdl,dong2022stacked}, we train the model on VG which includes the top 150 most frequent object classes and 50 predicate classes of VG.

\noindent\textbf{Tasks} On the VG dataset, performance evaluation of SGG models is typically based on three tasks, including Predicate Classification(PredCls), Scene Graph Classification (SGCls) and Scene Graph Detection (SGDet). PredCls requires predicting the relationship between a pair of objects when given bounding boxes and object labels. SGCls requires predicting both the object category and predicate category given the ground-truth object bounding boxes. SGDet has the most challenging setting, requiring predicting object bounding boxes, object categories, and their relationships.

\noindent\textbf{Metrics}
Following \cite{yan2020pcpl}, three different evaluation metrics are employed, including Recall (R@K), Mean recall (mR@K), and Mean. R@K measures the proportion of the correctly predicted relationships among the top K confident predictions\cite{lu2016visual}, which provides an evaluation of the model on the sample level. mR@K is obtained by calculating the average R@K score for each relationship\cite{tang2019learning}, reflecting the performance of the model on category level. Due to the influence of the long tailed distribution of training data, the SGG model usually has poor performance in recognizing tail categories with a small number of samples. So, both R and mR tend to reflect the performance of the model in certain categories. Mean computes the average of R@K and mR@K \cite{yan2020pcpl}, which provides a more comprehensive evaluation of the SGG model.

\subsection{Implementation Details}
We implemented the proposed method using PyTorch trained the model with a batch size of 12 on a NVIDIA GeForce RTX 3090 GPU. The Faster R-CNN \cite{ren2015faster} with ResXNet-101-FPN \cite{he2016deep} is employed as an object detector, and it remains frozen during training. The optimizer is SGD with a learning rate set to 0.0025. During the validation phase, the approach \cite{tang2019learning} is followed, introducing a validation set of 5,000 samples.

\subsection{Peformance Comparison}

Tab. \ref{tab1} shows the comparison of our method with the state-of-the-art models on the VG dataset. According to whether it can be used as a plugin to combine with other methods, the comparison methods are divided into two categories, model-specific methods and model-agnostic methods. The proposed EPD is a kind of model-agnostic method. For the model-agnostic methods, two robust SGG baselines, Motifs\cite{zellers2018neural} and VCTree\cite{tang2019learning}, are used for integration.

In Tab. \ref{tab1}, the regular models without debiasing strategies achieve the best performance on R@K, the proposed EPD performs best on Mean except for the SGCls task, and performs best on mR@K except for PredCls. The results demonstrate that the proposed EPD can effectively improve the prediction accuracy of low-frequency predicates without significantly reducing the prediction accuracy of high-frequency predicates.

Concretely, when using Motifs as the baseline, the proposed EPD outperforms the baseline and the other model-agnostic methods in terms of mR@K and Mean across all three tasks. Compared with the baseline Motifs, the other model-agnostic methods can improve mR@K but can not guarantee the improvement in Mean, such as the performance of Inf and GCL on the PredCls task.

When using VCTree as the baseline, the proposed EPD achieves
a competitive result. Specifically, EPD obtains the second Mean performance on PredCls and the third-best Mean performance on SGCls and SGDet, which is lower than Inf and DKBL. We analyzed the reason that they employed the triplet prior to the dataset, which receives a higher performance on R@K, further resulting in a higher Mean performance. However, our EPD performs better on mR@K, which indicates the more balanced performance of EPD among all predicates.

\subsection{Ablation Experiments}
\begin{table}[!htbp]
	\centering
	{
		\caption{Ablation results on EPD. }
		\begin{tabular}{l|ccc}
			\hline
			\multicolumn{1}{c|}{Settings} & R@50/100  & mR@50/100 & Mean \\ \hline
			baseline(Motifs)                    & 65.3/67.2 & 14.9/16.3 & 40.9 \\
			single-decoder+$\mathcal{L^{\prime}}_{all}$  & 48.1/50.6 & 28.5/30.3 & 39.4       \\
			multiple-decoder+$\mathcal{L^{\prime}}_{all}$          & 54.1/56.0 & 36.3/38.8 & 46.3 \\ \hline
		\end{tabular}
		\label{tab2}
	}
\end{table}
In this section, experiments are conducted for EPD, using one or three decoders in EPD, different structures of EPD, and different training data configurations for decoders in EPD. All ablation experiments are conducted on the PredCls task and use Motifs \cite{zellers2018neural} as the baseline.

\noindent\textbf{Ablation for the structure of EPD}
Ensemble predicate decoding is implemented through a multiple-decoder architecture and the loss $\mathcal{L}_{all}$. According to Eq. \ref{eq8}, the loss $\mathcal{L}_{all}$ is the weighted sum of $\mathcal{L}_{md}, \mathcal{L}_{ad_1}, \mathcal{L}_{ad_2}$ and $\mathcal{L}_{agg}$, which appears to have the same form of loss as the traditional re-weighting model. In order to verify the difference between EPD and traditional re-weighting models, we ablated the contribution of multiple-decoder and weighted loss to EPD. The ablation experiment is conducted using a single-decoder structure ($\mathcal{MD}$) and a multiple-decoder structure ($\mathcal{MD}$+$\mathcal{AD}_1$+$\mathcal{AD}_2$) respectively, while maintaining the same loss $\mathcal{L}_{all}$.

Denote the decoder's output in the single-decoder setting as $z$, then the $z_{md}$, $z_{ad_1}$, and $z_{ad_2}$ in $\mathcal{L}_{all}$ equals $z$. Define $\mathcal{L}_{N_1}$, $\mathcal{L}_{N_2}$, and $\mathcal{L}_{N_3}$ as the losses calculated for the head, body, and tail category predicates, respectively. Then, $\mathcal{L}_{md}$ can be expressed as $\mathcal{L}_{md} = \mathcal{L}_{N_1} + \mathcal{L}_{N_2} + \mathcal{L}_{N_3}$. Similarly, we can obtain the expression $\mathcal{L}_{ad_1} = \mathcal{L}_{N_2} + \mathcal{L}_{N_3}$, $\mathcal{L}_{ad_2} = \mathcal{L}_{N_3}$ and $\mathcal{L}_{agg} = \mathcal{L}_{N_1} + \mathcal{L}_{N_2} + \mathcal{L}_{N_3}$. Finally, the total loss $\mathcal{L^{\prime}}_{all}$ in the single-decoder setting is expressed as:

\begin{equation}
	\begin{split}
		\mathcal{L^{\prime}}_{all} &= \mathcal{L}_{N_1} + (1+\alpha)(1-\gamma)\mathcal{L}_{N_2} \\
		&\quad+ (1+\alpha+\beta)(1-\gamma)\mathcal{L}_{N_3}
	\end{split}
\end{equation}

The ablation results are shown in Tab. \ref{tab2}. Compared to the baseline, the single-decoder setting improves mR@K and shows a slight decrease in the Mean metric, indicating that the re-weighting can improve the prediction accuracy of low-frequency predicates. When using a multiple-decoder setting, mR@K and Mean are significantly improved, indicating that the combination with multiple-decoder setting can further improve the prediction accuracy of low-frequency predicates.

\noindent\textbf{Ablation for components of decoder}
\begin{table}[ht]
	\centering
	\caption{Ablation results in components of the decoder in EPD. ``Y'' indicates that the parameters of $F_{p\_d}$ in different decoders are shared, while ``N'' means independent. ``w.'' indicates that the BN layers $B_x$ is included in the decoders of EPD, while ``w.o.'' means that $B_x$ is not adopted.  }
	\begin{tabular}{cc|ccc}
		\hline
		$F_{p\_d}$ & $B_x$        & R@50/100  & mR@50/100 & Mean \\ \hline
		Y       & w.           & 54.1/56.0 & 36.3/38.8 & 46.3 \\
		N       & w.           & 49.9/52.0 & 35.0/37.3 & 43.5 \\
		Y       & w.o.    & 47.7/50.0 & 30.3/32.0 & 40.0  \\
		N       & w.o.           & 51.1/52.9 & 30.9/32.8 & 41.9 \\\hline
	\end{tabular}
	\label{tab3}
\end{table}
As shown in Fig. \ref{figure2}, each decoder in EPD consists of two linear layers and batch normalization(BN) layers. The linear layer parameters of the three decoders are shared, while the BN layer parameters are independent. In order to explore the role of BN layers and the impact of parameter sharing in linear layers, we conducted ablation experiments on the linear layer $F_{p\_d}$ and BN layers $B_d (d\in\{md, ad_1, ad_2\})$  of the decoder in EPD.

The ablation results are shown in Tab. \ref{tab3}. Comparing the first two rows and the last two rows, when the decoder does not include the BN layers, both mR@K and Mean decrease, meaning that independent BN is helpful for learning different subtasks. Comparing the first row with the second row, when the parameters of $F_{p\_d}$ are independent, both mR@K and Mean decrease, meaning that sharing initial features is important for the collaboration of the three decoders. Comparing the third row with the other rows, mR@K and Mean are the lowest because the model is a single-decoder structure when the parameters of $F_{p\_d}$ are shared and the BN layers are not included, while in the other cases, the model is multiple-decoder structure due to the independence of $F_{p\_d}$ or inclusion of the BN layers. This trend is consistent with the experimental results in Tab. \ref{tab2}, that is, the performance of a single-decoder is lower than that of a multiple-decoder.

\begin{figure*}
	\centering
	\includegraphics[scale=0.36]{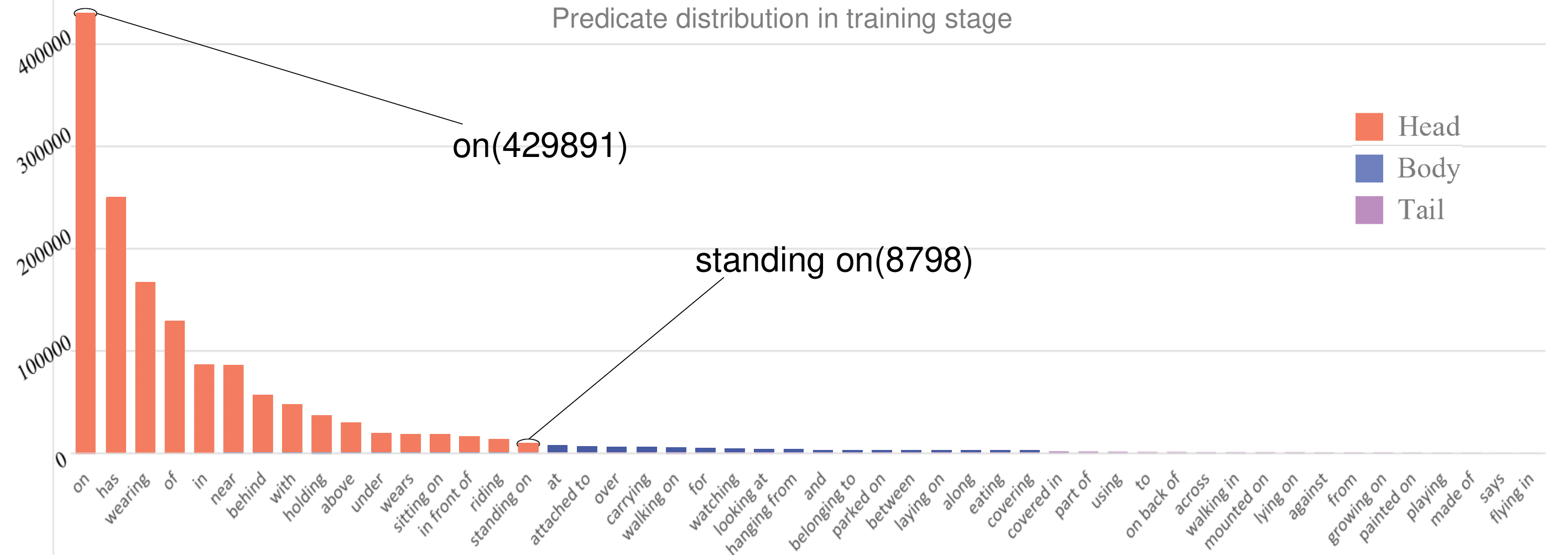}
	\caption{The predicate distribution in the training subsets ${N_1}$, ${N_2}$ and ${N_3}$ obtained by the average partitioning method, i.e., 16:17:17 (head:body:tail). The predicate samples used for statistics are from the VG dataset.}
	
	\label{figure3}
\end{figure*}

\noindent\textbf{Ablation of training data configurations}
In order to verify the impact of the training set configuration on the ensemble predicate decoding, we trained the model based on different data configurations. The experiment consists of three different configurations. Three decoders are trained on three disjoint subsets, i.e., main decoder $\mathcal{MD}$, auxiliary decoders $\mathcal{AD}_1$ and $\mathcal{AD}_2$ are assigned with the subsets $N_1$, $N_2$, and $N_3$ respectively. The main decoder $\mathcal{MD}$ is trained on the complete set of training samples ${N_1}\cup{N_2}\cup{N_3}$, and the auxiliary decoders $\mathcal{AD}_1$ and $\mathcal{AD}_2$ are trained on two disjoint subsets, i.e., $N_2$ and $N_3$ are assigned to $\mathcal{AD}_1$ and $\mathcal{AD}_2$  respectively. Three decoders are trained on three intersecting subsets, i.e., The main decoder $\mathcal{MD}$ is trained on ${N_1}\cup{N_2}\cup{N_3}$, the auxiliary decoder $\mathcal{AD}_1$ is trained on ${N_2}\cup{N_3}$, and the auxiliary decoder $\mathcal{AD}_2$ is trained on ${N_3}$. The latter two configurations allocate more decoders to predicates with lower frequencies.
\begin{table}[!htbp]
	\centering
	\caption{Ablation results on training data configuration. }
	\resizebox{0.47\textwidth}{!}{ 
		
		\begin{tabular}{c|c|c| c c c}
			\hline
			$\mathcal{MD}$&$\mathcal{AD}_1$&$\mathcal{AD}_2$&R@50/100&mR@50/100&Mean\\
			\hline
			$N_1$&$N_2$&$N_3$&59.9/61.5&22.0/24.1&41.9\\
			
			${N_1}\cup{N_2}\cup{N_3}$&${N_2}$&${N_3}$&52.2/54.1&34.8/37.2&44.6\\
			
			${N_1}\cup{N_2}\cup{N_3}$&${N_2}\cup{N_3}$&${N_3}$&54.1/56.0&36.3/38.8&46.3\\
			\hline
		\end{tabular}
		\label{tab4}
	}
\end{table}

The ablation results are shown in Tab. \ref{tab4}. When the three decoders are assigned with disjoint subsets, the mR@K and Mean are the lowest, as shown in the first row. When the training data assigned to decoders has an intersection, mR@K and Mean are improved. The predicates in the intersection of training data from different decoders belong to body predicate set ${N_2}$ or tail predicate set ${N_3}$, which have lower frequency and more informativeness compared to the predicates in the head predicate set ${N_1}$.
According to the viewpoint of Goel et al., \cite{goel2022not}, the prediction for the predicates with more informativeness requires deliberate reasoning. The configurations shown in the last two rows in Tab. \ref{tab4} allow the more informative samples to receive more attention, which makes subsequent inferences of such predicates more thoughtful.
\begin{table}[!htbp]
	\centering
	{
		\caption{Performance of EPD with different partition for ${N_1}$, ${N_2}$ and ${N_3}$. }
		\begin{tabular}{ccc|ccc}
			\hline
			\multicolumn{3}{c|}{Cardinality} & \multicolumn{3}{c}{Metric} \\
			$N_1$        & $N_2$        & $N_3$       & R@50/100     & mR@50/100   & Mean  \\ \hline
			16         & 17         & 17        & 57.4/59.2    & 33.3/35.8   & 46.4   \\
			10         & 10         & 30        & 54.9/56.7    & 35.2/37.6   & 46.1    \\
			5          & 10         & 35        & 54.1/56.0    & 36.3/38.8   & 46.3    \\
			5          & 15         & 30        & 52.5/54.5    & 35.8/38.1   & 45.2    \\ \hline
		\end{tabular}
		\label{tab5}
	}
\end{table}

\subsection{Hyperparameter Selection Experiments}
In this section, experiments are conducted to select the optimal hyperparameter values. All experiments are conducted on the PredCls task and use Motifs \cite{zellers2018neural} as a baseline.

\noindent \textbf{The cardinality of ${N_1}$, ${N_2}$ and ${N_3}$ } Previous works \cite{li2022devil,li2021bipartite} set ${N_1}$, ${N_2}$ and ${N_3}$ include 16, 17, and 17 categories, respectively. The partitioning process first arranges the predicate categories in descending order based on the frequency of their associated samples in the dataset, and then evenly divides the predicate categories in the sequence. As shown in Fig. \ref{figure3}, when the predicates are evenly divided, predicate pairs with similar semantics, such as \textit{standing on} and \textit{on}, are grouped into the same subset. This situation will increase the difficulty of the task faced by the decoder. In addition, due to the large variance of predicate frequency, the ${N_1}$ obtained by averaging based on the number of categories includes predicate categories with lower frequencies. The idea of EPD is to make low-frequency predicates participate more in the training process than high-frequency predicates, so the partition containing fewer low-frequency predicate categories in ${N_1}$ is more advantageous for EPD.

\begin{table}[!htbp]
	\centering
	{
		\caption{Coefficients experiments on the loss function. The "Head," "Body," and "Tail" represent the averaged recall rates of predicates within their respective groups. The results are obtained with Motifs as the baseline, in the PredCls task.}
		\begin{tabular}{*{3}{>{\centering\arraybackslash}p{0.6cm}}*{4}{>{\centering\arraybackslash}p{0.8cm}}}
			\hline
			
			\multicolumn{3}{c|}{Weights} & \multicolumn{4}{c}{mR@100}    \\
			\hline
			\multicolumn{1}{c}{$\alpha$} & \multicolumn{1}{c}{$\beta$} & \multicolumn{1}{c|}{$\gamma$}  & Head & Body & Tail & All \\
			\hline
			8 & 10 & \multicolumn{1}{l|}{0.02} & 54.0 & 24.9 & 35.7 & 36.4 \\
			8 & 10 & \multicolumn{1}{l|}{0.005} & 51.8 & 33.9 & 35.4 & 37.6 \\
			8 & 10 & \multicolumn{1}{l|}{0.01} & 53.8 & 30.4 & 37.7 & 38.8 \\
			2 & 3 & \multicolumn{1}{l|}{0.01} & 57.5 & 40.3 & 27.4 & 33.6 \\
			\hline
		\end{tabular}
		\label{tab6}
	}
	
\end{table}
Based on different partition principles, we obtained multiple sets of training subsets and used each set of training subsets to train EPD. The corresponding model performance is listed in Tab. \ref{tab5}. Comparing the first row and the last two rows, when ${N_1}$ contains fewer categories, mR@K of the model is higher. Comparing the first row and the third row, when ${N_1}$ contains fewer categories, the model has a higher mR@K and a similar Mean. We ultimately set the number of samples in ${N_1}$, ${N_2}$ and ${N_3}$ to 5, 10, and 35, respectively.

\noindent\textbf{Weights in Loss Function:}
In Eq. \ref{eq8}, there are three weights $\alpha$, $\beta$ and $\gamma$ in the total loss $\mathcal{L}_{all}$. $\alpha$ and $\beta$ are used to balance auxiliary decoders and the main decoder, $\gamma$ is utilized to balance individual predictions and the aggregated prediction. We conduct experiments with various settings for $\alpha$, $\beta$ and $\gamma$. The experimental results are shown in Tab. \ref{tab6}. When $\gamma$ decreases, the recall rates of the three groups become more balanced. When the value of $\alpha \slash \beta$ is less than 1, mR@100 is higher. As indicated in the third row, the model achieves the best mR@K while $\alpha=8$, $\beta=10$, and $\gamma=0.01$.

\begin{table}[!htbp]
	\centering
	{
		\caption{Coefficients experiments on the loss function. The "Head," "Body," and "Tail" represent the averaged recall rates of predicates within their respective groups. The results are obtained with Motifs as the baseline, in the PredCls task.}
		\begin{tabular}{*{3}{>{\centering\arraybackslash}p{0.6cm}}*{4}{>{\centering\arraybackslash}p{0.8cm}}}
			\hline
			
			\multicolumn{3}{c|}{Weights} & \multicolumn{4}{c}{mR@100}    \\
			\hline
			\multicolumn{1}{c}{$\alpha$} & \multicolumn{1}{c}{$\beta$} & \multicolumn{1}{c|}{$\gamma$}  & Head & Body & Tail & All \\
			\hline
			8 & 10 & \multicolumn{1}{l|}{0.02} & 54.0 & 24.9 & 35.7 & 36.4 \\
			8 & 10 & \multicolumn{1}{l|}{0.005} & 51.8 & 33.9 & 35.4 & 37.6 \\
			8 & 10 & \multicolumn{1}{l|}{0.01} & 53.8 & 30.4 & 37.7 & 38.8 \\
			2 & 3 & \multicolumn{1}{l|}{0.01} & 57.5 & 40.3 & 27.4 & 33.6 \\
			\hline
		\end{tabular}
		\label{tab6}
	}
	
\end{table}
\noindent\textbf{Weights in the aggregated prediction}
According to Eq. \ref{eq7}, the aggregated prediction is the weighted sum of the individual prediction. Compared with the main decoder $\mathcal{MD}$, auxiliary decoders $\mathcal{AD}_1$ and $\mathcal{AD}_2$ are trained on a relatively limited set of predicate categories. Thus, $\mathcal{AD}_1$ and $\mathcal{AD}_2$ should play a supporting role in serving as supplements to assist the main decoder $\mathcal{MD}$ in making predictions. In terms of $\mathcal{AD}_1$ and $\mathcal{AD}_2$, focusing more on low-frequency samples helps the model reduce bias effects. We conduct experiments with various settings for the weights $\lambda_{md}$, $\lambda_{ad_1}$ and $\lambda_{ad_2}$. The experimental results are shown in Tab. \ref{tab7}. $\lambda_{md}$ is always given the maximum value, while $\lambda_{ad_1}$ and $\lambda_{ad_2}$ are given smaller values. As indicated in the third row, the model achieves the best mR@K while $\lambda_{md}=0.4$, $\lambda_{ad_1}=0.2$ and $\lambda_{ad_2}=0.4$.
\begin{figure*}
	\centering
	\includegraphics[scale=0.75]{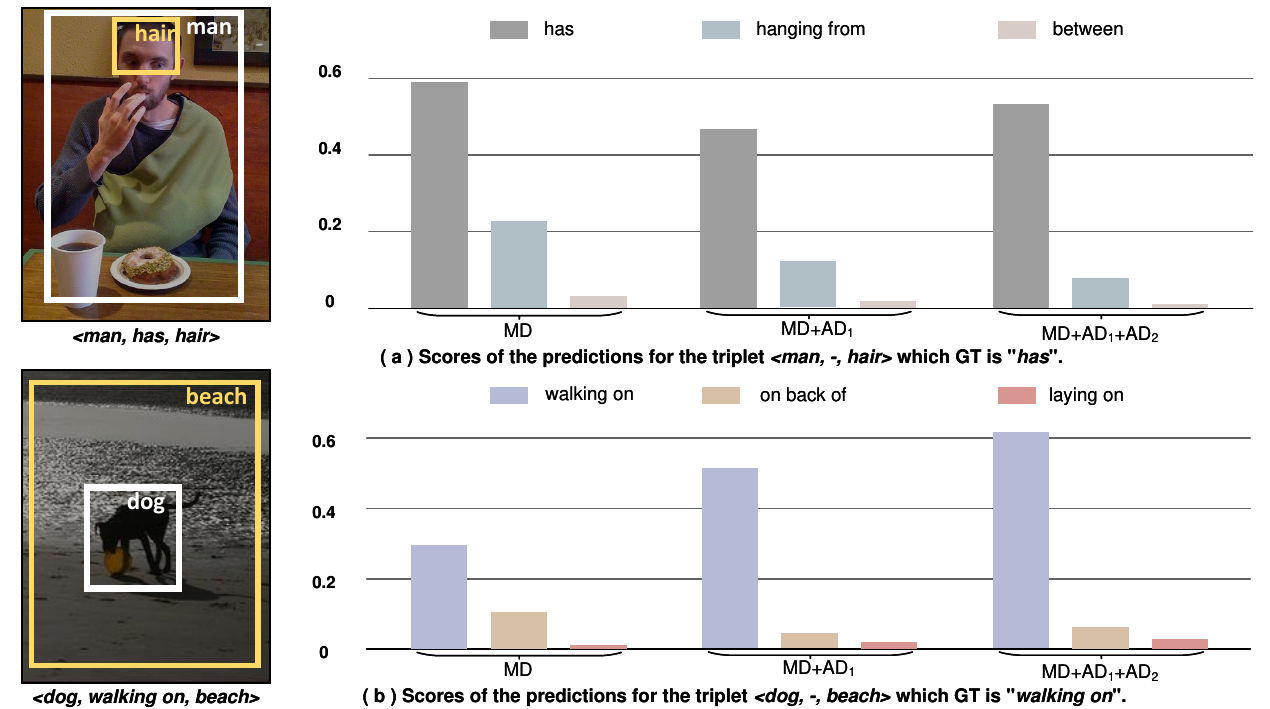}
	\caption{Visualization of collaborative scoring with multiple decoders, using Motifs with EPD. The head predicate triplet $\langle \textit{man}, \textit{has}, \textit{hair} \rangle$ and the body predicate triplet $\langle \textit{dog}, \textit{walking on}, \textit{beach} \rangle$ are taken as examples. For each triplet, the scores given by EPD using different decoder combinations are shown, including: main decoder only; main decoder and auxiliary decoder 1; main decoder, auxiliary decoder 1, and auxiliary decoder 2. }
	\label{fig4}
\end{figure*}
\begin{table}[!htbp]
	\centering
	{
		\caption{Performance with different settings for $\lambda_{md}$, $\lambda_{ad_1}$ and $\lambda_{ad_2}$ in $z_{sum}$.}
		\begin{tabular}{llllll}
			\hline
			$\lambda_{md}$ & $\lambda_{ad_1}$    & $\lambda_{ad_2}$ & R@50/100  & mR@50/100 & Mean \\ \hline
			0.4 & 0.3 & 0.3 & 43.1/45.3 & 37.8/40.3 & 41.6 \\
			0.5 & 0.3 & 0.2 & 48.4/50.8 & 34.4/36.8 & 42.6 \\
			0.4 & 0.2 & 0.4 & 46.5/48.7 & 38.2/40.9 & 43.5 \\
			0.5 & 0.2 & 0.3 & 54.1/56.0 & 36.3/38.8 & 46.3 \\ \hline
		\end{tabular}
		\label{tab7}
	}
	
\end{table}
\subsection{Visualization}
In Fig. \ref{figure1}, for the tail predicate triplet $\langle \textit{bird}, \textit{--}, \textit{banana} \rangle$, the prediction becomes correct by combining auxiliary decoders to continuously improve the rating of $\langle \textit{standing on} \rangle$. Do the auxiliary decoders have a negative impact? Therefore, we conducted experiments of visualizing the collaborative scoring with multiple decoders for the triplets which ground truths are head and body predicates.
The visualization results are shown in Fig. \ref{fig4}. The instance in the top row is a head predicate $\langle \textit{man}, \textit{--}, \textit{hair} \rangle$ which ground truth is \textit{has}. When only including $\mathcal{MD}$, EPD gives the correct prediction. When auxiliary decoders $\mathcal{AD}_1$ and $\mathcal{AD}_2$ are combined progressively, the prediction still remains correct. In the bottom row, the prediction of the body predicate \textit{walking on} is also the same situation. The reason is that the auxiliary decoder $\mathcal{AD}_2$ can make reasonable judgments on learned tail predicates, such as giving \textit{hanging from} and \textit{on back of} a lower score, thereby reducing the scores of non-ground truth predicates.

\section{Conclusion}
In this paper, we analyzed the influencing factors beyond the long tail of the dataset that led to biased predictions in the SGG model. Based on the analysis, we introduce a model-agnostic strategy called Ensemble Predicate Decoding (EPD). EPD enhances the model's decoding capability for infrequent categories by training multiple decoders learned on different subtasks. Experimental results on the VG dataset demonstrate the excellent performance of our method across different baseline models.
\section*{Acknowledgements}
This work was supported by the National KeyR\&D Program of China (No. 2021ZD0111902), National Natural Science Foundation of China(U21B2038, 62376014, 62172022), and Foundation for China universityIndustry-university Research Innovation(No.2021JQR023)
\bibliographystyle{cas-model2-names}
\bibliography{cas-refs}

\begin{thebibliography}{37}
\expandafter\ifx\csname natexlab\endcsname\relax\def\natexlab#1{#1}\fi
\providecommand{\url}[1]{\texttt{#1}}
\providecommand{\href}[2]{#2}
\providecommand{\path}[1]{#1}
\providecommand{\DOIprefix}{doi:}
\providecommand{\ArXivprefix}{arXiv:}
\providecommand{\URLprefix}{URL: }
\providecommand{\Pubmedprefix}{pmid:}
\providecommand{\doi}[1]{\href{http://dx.doi.org/#1}{\path{#1}}}
\providecommand{\Pubmed}[1]{\href{pmid:#1}{\path{#1}}}
\providecommand{\bibinfo}[2]{#2}
\ifx\xfnm\relax \def\xfnm[#1]{\unskip,\space#1}\fi
\bibitem[{Biswas and Ji(2023)}]{biswas2023probabilistic}
\bibinfo{author}{Biswas, B.A.}, \bibinfo{author}{Ji, Q.}, \bibinfo{year}{2023}.
\newblock \bibinfo{title}{Probabilistic debiasing of scene graphs}, in:
  \bibinfo{booktitle}{Proceedings of the IEEE/CVF Conference on Computer Vision
  and Pattern Recognition}, pp. \bibinfo{pages}{10429--10438}.
\bibitem[{Chen et~al.(2023)Chen, Li, Luo and Xiao}]{chen2023addressing}
\bibinfo{author}{Chen, G.}, \bibinfo{author}{Li, L.}, \bibinfo{author}{Luo,
  Y.}, \bibinfo{author}{Xiao, J.}, \bibinfo{year}{2023}.
\newblock \bibinfo{title}{Addressing predicate overlap in scene graph
  generation with semantic granularity controller}, in:
  \bibinfo{booktitle}{2023 IEEE International Conference on Multimedia and
  Expo}, \bibinfo{organization}{IEEE}. pp. \bibinfo{pages}{78--83}.
\bibitem[{Chen et~al.(2020)Chen, Jin, Wang and Wu}]{chen2020say}
\bibinfo{author}{Chen, S.}, \bibinfo{author}{Jin, Q.}, \bibinfo{author}{Wang,
  P.}, \bibinfo{author}{Wu, Q.}, \bibinfo{year}{2020}.
\newblock \bibinfo{title}{Say as you wish: Fine-grained control of image
  caption generation with abstract scene graphs}, in:
  \bibinfo{booktitle}{Proceedings of the IEEE/CVF Conference on Computer Vision
  and Pattern Recognition}, pp. \bibinfo{pages}{9962--9971}.
\bibitem[{Cui et~al.(2022)Cui, Liu, Tian, Zhong and Jia}]{cui2022reslt}
\bibinfo{author}{Cui, J.}, \bibinfo{author}{Liu, S.}, \bibinfo{author}{Tian,
  Z.}, \bibinfo{author}{Zhong, Z.}, \bibinfo{author}{Jia, J.},
  \bibinfo{year}{2022}.
\newblock \bibinfo{title}{Reslt: Residual learning for long-tailed
  recognition}.
\newblock \bibinfo{journal}{IEEE Transactions on Pattern Analysis and Machine
  Intelligence} \bibinfo{volume}{45}, \bibinfo{pages}{3695--3706}.
\bibitem[{Dong et~al.(2022)Dong, Gan, Song, Wu, Cheng and
  Nie}]{dong2022stacked}
\bibinfo{author}{Dong, X.}, \bibinfo{author}{Gan, T.}, \bibinfo{author}{Song,
  X.}, \bibinfo{author}{Wu, J.}, \bibinfo{author}{Cheng, Y.},
  \bibinfo{author}{Nie, L.}, \bibinfo{year}{2022}.
\newblock \bibinfo{title}{Stacked hybrid-attention and group collaborative
  learning for unbiased scene graph generation}, in:
  \bibinfo{booktitle}{Proceedings of the IEEE/CVF Conference on Computer Vision
  and Pattern Recognition}, pp. \bibinfo{pages}{19427--19436}.
\bibitem[{Garc{\'\i}a-Pedrajas(2009)}]{garcia2009constructing}
\bibinfo{author}{Garc{\'\i}a-Pedrajas, N.}, \bibinfo{year}{2009}.
\newblock \bibinfo{title}{Constructing ensembles of classifiers by means of
  weighted instance selection}.
\newblock \bibinfo{journal}{IEEE Transactions on Neural Networks}
  \bibinfo{volume}{20}, \bibinfo{pages}{258--277}.
\bibitem[{Geiselhart et~al.(2021)Geiselhart, Elkelesh, Ebada, Cammerer and ten
  Brink}]{geiselhart2021automorphism}
\bibinfo{author}{Geiselhart, M.}, \bibinfo{author}{Elkelesh, A.},
  \bibinfo{author}{Ebada, M.}, \bibinfo{author}{Cammerer, S.},
  \bibinfo{author}{ten Brink, S.}, \bibinfo{year}{2021}.
\newblock \bibinfo{title}{Automorphism ensemble decoding of reed--muller
  codes}.
\newblock \bibinfo{journal}{IEEE Transactions on Communications}
  \bibinfo{volume}{69}, \bibinfo{pages}{6424--6438}.
\bibitem[{Goel et~al.(2022)Goel, Fernando, Keller and Bilen}]{goel2022not}
\bibinfo{author}{Goel, A.}, \bibinfo{author}{Fernando, B.},
  \bibinfo{author}{Keller, F.}, \bibinfo{author}{Bilen, H.},
  \bibinfo{year}{2022}.
\newblock \bibinfo{title}{Not all relations are equal: Mining informative
  labels for scene graph generation}, in: \bibinfo{booktitle}{Proceedings of
  the IEEE/CVF Conference on Computer Vision and Pattern Recognition}, pp.
  \bibinfo{pages}{15596--15606}.
\bibitem[{Han et~al.(2022)Han, Zhuo, Zhang, Huang, Zha, Zhang and
  Kankanhalli}]{han2022one}
\bibinfo{author}{Han, Y.}, \bibinfo{author}{Zhuo, T.}, \bibinfo{author}{Zhang,
  P.}, \bibinfo{author}{Huang, W.}, \bibinfo{author}{Zha, Y.},
  \bibinfo{author}{Zhang, Y.}, \bibinfo{author}{Kankanhalli, M.},
  \bibinfo{year}{2022}.
\newblock \bibinfo{title}{One-shot video graph generation for explainable
  action reasoning}.
\newblock \bibinfo{journal}{Neurocomputing} \bibinfo{volume}{488},
  \bibinfo{pages}{212--225}.
\bibitem[{He et~al.(2016)He, Zhang, Ren and Sun}]{he2016deep}
\bibinfo{author}{He, K.}, \bibinfo{author}{Zhang, X.}, \bibinfo{author}{Ren,
  S.}, \bibinfo{author}{Sun, J.}, \bibinfo{year}{2016}.
\newblock \bibinfo{title}{Deep residual learning for image recognition}, in:
  \bibinfo{booktitle}{Proceedings of the IEEE conference on computer vision and
  pattern recognition}, pp. \bibinfo{pages}{770--778}.
\bibitem[{Herzig et~al.(2020)Herzig, Bar, Xu, Chechik, Darrell and
  Globerson}]{herzig2020learning}
\bibinfo{author}{Herzig, R.}, \bibinfo{author}{Bar, A.}, \bibinfo{author}{Xu,
  H.}, \bibinfo{author}{Chechik, G.}, \bibinfo{author}{Darrell, T.},
  \bibinfo{author}{Globerson, A.}, \bibinfo{year}{2020}.
\newblock \bibinfo{title}{Learning canonical representations for scene graph to
  image generation}, in: \bibinfo{booktitle}{European Conference on Computer
  Vision}, pp. \bibinfo{pages}{210--227}.
\bibitem[{Johnson et~al.(2015)Johnson, Krishna, Stark, Li, Shamma, Bernstein
  and Fei-Fei}]{johnson2015image}
\bibinfo{author}{Johnson, J.}, \bibinfo{author}{Krishna, R.},
  \bibinfo{author}{Stark, M.}, \bibinfo{author}{Li, L.J.},
  \bibinfo{author}{Shamma, D.}, \bibinfo{author}{Bernstein, M.},
  \bibinfo{author}{Fei-Fei, L.}, \bibinfo{year}{2015}.
\newblock \bibinfo{title}{Image retrieval using scene graphs}, in:
  \bibinfo{booktitle}{Proceedings of the IEEE/CVF Conference on Computer Vision
  and Pattern Recognition}, pp. \bibinfo{pages}{3668--3678}.
\bibitem[{Krishna et~al.(2017)Krishna, Zhu, Groth, Johnson, Hata, Kravitz,
  Chen, Kalantidis, Li, Shamma et~al.}]{krishna2017visual}
\bibinfo{author}{Krishna, R.}, \bibinfo{author}{Zhu, Y.},
  \bibinfo{author}{Groth, O.}, \bibinfo{author}{Johnson, J.},
  \bibinfo{author}{Hata, K.}, \bibinfo{author}{Kravitz, J.},
  \bibinfo{author}{Chen, S.}, \bibinfo{author}{Kalantidis, Y.},
  \bibinfo{author}{Li, L.J.}, \bibinfo{author}{Shamma, D.A.}, et~al.,
  \bibinfo{year}{2017}.
\newblock \bibinfo{title}{Visual genome: Connecting language and vision using
  crowdsourced dense image annotations}.
\newblock \bibinfo{journal}{International Journal of Computer Vision}
  \bibinfo{volume}{123}, \bibinfo{pages}{32--73}.
\bibitem[{Li et~al.(2024)Li, Zhu, Zhang, Jiang, Dang, Hou, Shen, Zhao, Shah and
  Bennamoun}]{li2024scene}
\bibinfo{author}{Li, H.}, \bibinfo{author}{Zhu, G.}, \bibinfo{author}{Zhang,
  L.}, \bibinfo{author}{Jiang, Y.}, \bibinfo{author}{Dang, Y.},
  \bibinfo{author}{Hou, H.}, \bibinfo{author}{Shen, P.}, \bibinfo{author}{Zhao,
  X.}, \bibinfo{author}{Shah, S.A.A.}, \bibinfo{author}{Bennamoun, M.},
  \bibinfo{year}{2024}.
\newblock \bibinfo{title}{Scene graph generation: A comprehensive survey}.
\newblock \bibinfo{journal}{Neurocomputing} \bibinfo{volume}{566},
  \bibinfo{pages}{127052}.
\bibitem[{Li et~al.(2022a)Li, Chen, Huang, Zhang, Zhang and Xiao}]{li2022devil}
\bibinfo{author}{Li, L.}, \bibinfo{author}{Chen, L.}, \bibinfo{author}{Huang,
  Y.}, \bibinfo{author}{Zhang, Z.}, \bibinfo{author}{Zhang, S.},
  \bibinfo{author}{Xiao, J.}, \bibinfo{year}{2022}a.
\newblock \bibinfo{title}{The devil is in the labels: Noisy label correction
  for robust scene graph generation}, in: \bibinfo{booktitle}{Proceedings of
  the IEEE/CVF Conference on Computer Vision and Pattern Recognition}, pp.
  \bibinfo{pages}{18869--18878}.
\bibitem[{Li et~al.(2023)Li, Xiao, Shi, Wang, Shao, Liu, Yang and
  Chen}]{li2023label}
\bibinfo{author}{Li, L.}, \bibinfo{author}{Xiao, J.}, \bibinfo{author}{Shi,
  H.}, \bibinfo{author}{Wang, W.}, \bibinfo{author}{Shao, J.},
  \bibinfo{author}{Liu, A.A.}, \bibinfo{author}{Yang, Y.},
  \bibinfo{author}{Chen, L.}, \bibinfo{year}{2023}.
\newblock \bibinfo{title}{Label semantic knowledge distillation for unbiased
  scene graph generation}.
\newblock \bibinfo{journal}{IEEE Transactions on Circuits and Systems for Video
  Technology} \bibinfo{volume}{34}, \bibinfo{pages}{195--206}.
\bibitem[{Li et~al.(2021)Li, Zhang, Wan and He}]{li2021bipartite}
\bibinfo{author}{Li, R.}, \bibinfo{author}{Zhang, S.}, \bibinfo{author}{Wan,
  B.}, \bibinfo{author}{He, X.}, \bibinfo{year}{2021}.
\newblock \bibinfo{title}{Bipartite graph network with adaptive message passing
  for unbiased scene graph generation}, in: \bibinfo{booktitle}{Proceedings of
  the IEEE/CVF Conference on Computer Vision and Pattern Recognition}, pp.
  \bibinfo{pages}{11109--11119}.
\bibitem[{Li et~al.(2022b)Li, Zhang, Bai, Zhao, Jiang and Yuan}]{li2022ppdl}
\bibinfo{author}{Li, W.}, \bibinfo{author}{Zhang, H.}, \bibinfo{author}{Bai,
  Q.}, \bibinfo{author}{Zhao, G.}, \bibinfo{author}{Jiang, N.},
  \bibinfo{author}{Yuan, X.}, \bibinfo{year}{2022}b.
\newblock \bibinfo{title}{Ppdl: Predicate probability distribution based loss
  for unbiased scene graph generation}, in: \bibinfo{booktitle}{Proceedings of
  the IEEE/CVF Conference on Computer Vision and Pattern Recognition}, pp.
  \bibinfo{pages}{19447--19456}.
\bibitem[{Lin et~al.(2020)Lin, Ding, Zeng and Tao}]{lin2020gps}
\bibinfo{author}{Lin, X.}, \bibinfo{author}{Ding, C.}, \bibinfo{author}{Zeng,
  J.}, \bibinfo{author}{Tao, D.}, \bibinfo{year}{2020}.
\newblock \bibinfo{title}{Gps-net: Graph property sensing network for scene
  graph generation}, in: \bibinfo{booktitle}{Proceedings of the IEEE/CVF
  Conference on Computer Vision and Pattern Recognition}, pp.
  \bibinfo{pages}{3746--3753}.
\bibitem[{Lu et~al.(2016a)Lu, Krishna, Bernstein and Fei-Fei}]{lu2016visual}
\bibinfo{author}{Lu, C.}, \bibinfo{author}{Krishna, R.},
  \bibinfo{author}{Bernstein, M.}, \bibinfo{author}{Fei-Fei, L.},
  \bibinfo{year}{2016}a.
\newblock \bibinfo{title}{Visual relationship detection with language priors},
  in: \bibinfo{booktitle}{European Conference on Computer Vision},
  \bibinfo{organization}{Springer}. pp. \bibinfo{pages}{852--869}.
\bibitem[{Lu et~al.(2016b)Lu, Yang, Batra and Parikh}]{lu2016hierarchical}
\bibinfo{author}{Lu, J.}, \bibinfo{author}{Yang, J.}, \bibinfo{author}{Batra,
  D.}, \bibinfo{author}{Parikh, D.}, \bibinfo{year}{2016}b.
\newblock \bibinfo{title}{Hierarchical question-image co-attention for visual
  question answering}, in: \bibinfo{booktitle}{Proceedings of the 30th
  International Conference on Neural Information Processing Systems}, p.
  \bibinfo{pages}{289–297}.
\bibitem[{Nguyen et~al.(2024)Nguyen, Nguyen and Luu}]{nguyen2024hig}
\bibinfo{author}{Nguyen, T.T.}, \bibinfo{author}{Nguyen, P.},
  \bibinfo{author}{Luu, K.}, \bibinfo{year}{2024}.
\newblock \bibinfo{title}{Hig: Hierarchical interlacement graph approach to
  scene graph generation in video understanding}, in:
  \bibinfo{booktitle}{Proceedings of the IEEE/CVF Conference on Computer Vision
  and Pattern Recognition}, pp. \bibinfo{pages}{18384--18394}.
\bibitem[{Peng et~al.(2024)Peng, Zhao, Chen, Wang, Liu, Liu and
  Lan}]{peng2024causality}
\bibinfo{author}{Peng, R.}, \bibinfo{author}{Zhao, C.}, \bibinfo{author}{Chen,
  X.}, \bibinfo{author}{Wang, Z.}, \bibinfo{author}{Liu, Y.},
  \bibinfo{author}{Liu, Y.}, \bibinfo{author}{Lan, X.}, \bibinfo{year}{2024}.
\newblock \bibinfo{title}{A causality guided loss for imbalanced learning in
  scene graph generation}.
\newblock \bibinfo{journal}{Neurocomputing} , \bibinfo{pages}{128042}.
\bibitem[{Polikar(2012)}]{polikar2012ensemble}
\bibinfo{author}{Polikar, R.}, \bibinfo{year}{2012}.
\newblock \bibinfo{title}{Ensemble learning}.
\newblock \bibinfo{journal}{Ensemble machine learning: Methods and
  applications} , \bibinfo{pages}{1--34}.
\bibitem[{Ren et~al.(2017)Ren, He, Girshick and Sun}]{ren2015faster}
\bibinfo{author}{Ren, S.}, \bibinfo{author}{He, K.}, \bibinfo{author}{Girshick,
  R.}, \bibinfo{author}{Sun, J.}, \bibinfo{year}{2017}.
\newblock \bibinfo{title}{Faster r-cnn: Towards real-time object detection with
  region proposal networks}.
\newblock \bibinfo{journal}{IEEE Transactions on Pattern Analysis and Machine
  Intelligence} \bibinfo{volume}{39}.
\bibitem[{Shen et~al.(2021)Shen, Yu, Ma and Kwok}]{shen2021time}
\bibinfo{author}{Shen, L.}, \bibinfo{author}{Yu, Z.}, \bibinfo{author}{Ma, Q.},
  \bibinfo{author}{Kwok, J.T.}, \bibinfo{year}{2021}.
\newblock \bibinfo{title}{Time series anomaly detection with multiresolution
  ensemble decoding}, in: \bibinfo{booktitle}{Proceedings of the AAAI
  Conference on Artificial Intelligence}, pp. \bibinfo{pages}{9567--9575}.
\bibitem[{Song et~al.(2021)Song, Chen, Wu and Jiang}]{song2021spatial}
\bibinfo{author}{Song, X.}, \bibinfo{author}{Chen, J.}, \bibinfo{author}{Wu,
  Z.}, \bibinfo{author}{Jiang, Y.G.}, \bibinfo{year}{2021}.
\newblock \bibinfo{title}{Spatial-temporal graphs for cross-modal text2video
  retrieval}.
\newblock \bibinfo{journal}{IEEE Transactions on Multimedia}
  \bibinfo{volume}{24}, \bibinfo{pages}{2914--2923}.
\bibitem[{Tang et~al.(2020)Tang, Niu, Huang, Shi and Zhang}]{tang2020unbiased}
\bibinfo{author}{Tang, K.}, \bibinfo{author}{Niu, Y.}, \bibinfo{author}{Huang,
  J.}, \bibinfo{author}{Shi, J.}, \bibinfo{author}{Zhang, H.},
  \bibinfo{year}{2020}.
\newblock \bibinfo{title}{Unbiased scene graph generation from biased
  training}, in: \bibinfo{booktitle}{Proceedings of the IEEE/CVF conference on
  computer vision and pattern recognition}, pp. \bibinfo{pages}{3716--3725}.
\bibitem[{Tang et~al.(2019)Tang, Zhang, Wu, Luo and Liu}]{tang2019learning}
\bibinfo{author}{Tang, K.}, \bibinfo{author}{Zhang, H.}, \bibinfo{author}{Wu,
  B.}, \bibinfo{author}{Luo, W.}, \bibinfo{author}{Liu, W.},
  \bibinfo{year}{2019}.
\newblock \bibinfo{title}{Learning to compose dynamic tree structures for
  visual contexts}, in: \bibinfo{booktitle}{Proceedings of the IEEE/CVF
  Conference on Computer Vision and Pattern Recognition}, pp.
  \bibinfo{pages}{6619--6628}.
\bibitem[{Tong et~al.(2023)Tong, Wang and Jing}]{tong2023alleviating}
\bibinfo{author}{Tong, X.}, \bibinfo{author}{Wang, R.}, \bibinfo{author}{Jing,
  L.}, \bibinfo{year}{2023}.
\newblock \bibinfo{title}{Alleviating training bias with less cost via
  multi-expert de-biasing method in scene graph generation}, in:
  \bibinfo{booktitle}{Proceedings of the 1st International Workshop on
  Multimedia Content Generation and Evaluation: New Methods and Practice}, pp.
  \bibinfo{pages}{99--103}.
\bibitem[{Yan et~al.(2020)Yan, Shen, Jin, Huang, Jiang, Chen and
  Hua}]{yan2020pcpl}
\bibinfo{author}{Yan, S.}, \bibinfo{author}{Shen, C.}, \bibinfo{author}{Jin,
  Z.}, \bibinfo{author}{Huang, J.}, \bibinfo{author}{Jiang, R.},
  \bibinfo{author}{Chen, Y.}, \bibinfo{author}{Hua, X.S.},
  \bibinfo{year}{2020}.
\newblock \bibinfo{title}{Pcpl: Predicate-correlation perception learning for
  unbiased scene graph generation}, in: \bibinfo{booktitle}{Proceedings of the
  28th ACM International Conference on Multimedia}, pp.
  \bibinfo{pages}{265--273}.
\bibitem[{Yang et~al.(2021)Yang, Zhang, Zhang, Wu and
  Yang}]{yang2021probabilistic}
\bibinfo{author}{Yang, G.}, \bibinfo{author}{Zhang, J.},
  \bibinfo{author}{Zhang, Y.}, \bibinfo{author}{Wu, B.}, \bibinfo{author}{Yang,
  Y.}, \bibinfo{year}{2021}.
\newblock \bibinfo{title}{Probabilistic modeling of semantic ambiguity for
  scene graph generation}, in: \bibinfo{booktitle}{Proceedings of the IEEE/CVF
  Conference on Computer Vision and Pattern Recognition}, pp.
  \bibinfo{pages}{12527--12536}.
\bibitem[{Yoon et~al.(2023)Yoon, Kim, Moon and Park}]{yoon2023unbiased}
\bibinfo{author}{Yoon, K.}, \bibinfo{author}{Kim, K.}, \bibinfo{author}{Moon,
  J.}, \bibinfo{author}{Park, C.}, \bibinfo{year}{2023}.
\newblock \bibinfo{title}{Unbiased heterogeneous scene graph generation with
  relation-aware message passing neural network}, in:
  \bibinfo{booktitle}{Proceedings of the AAAI Conference on Artificial
  Intelligence}, pp. \bibinfo{pages}{3285--3294}.
\bibitem[{Yu et~al.(2021)Yu, Chai, Wang, Hu and Wu}]{yu2020cogtree}
\bibinfo{author}{Yu, J.}, \bibinfo{author}{Chai, Y.}, \bibinfo{author}{Wang,
  Y.}, \bibinfo{author}{Hu, Y.}, \bibinfo{author}{Wu, Q.},
  \bibinfo{year}{2021}.
\newblock \bibinfo{title}{Cogtree: Cognition tree loss for unbiased scene graph
  generation}, in: \bibinfo{booktitle}{Proceedings of the Thirtieth
  International Joint Conference on Artificial Intelligence}, pp.
  \bibinfo{pages}{1274--1280}.
\bibitem[{Zellers et~al.(2018)Zellers, Yatskar, Thomson and
  Choi}]{zellers2018neural}
\bibinfo{author}{Zellers, R.}, \bibinfo{author}{Yatskar, M.},
  \bibinfo{author}{Thomson, S.}, \bibinfo{author}{Choi, Y.},
  \bibinfo{year}{2018}.
\newblock \bibinfo{title}{Neural motifs: Scene graph parsing with global
  context}, in: \bibinfo{booktitle}{Proceedings of the IEEE/CVF Conference on
  Computer Vision and Pattern Recognition}, pp. \bibinfo{pages}{5831--5840}.
\bibitem[{Zeng et~al.(2021)Zeng, Gao, Lyu, Jing and Song}]{zeng2021conceptual}
\bibinfo{author}{Zeng, P.}, \bibinfo{author}{Gao, L.}, \bibinfo{author}{Lyu,
  X.}, \bibinfo{author}{Jing, S.}, \bibinfo{author}{Song, J.},
  \bibinfo{year}{2021}.
\newblock \bibinfo{title}{Conceptual and syntactical cross-modal alignment with
  cross-level consistency for image-text matching}, in:
  \bibinfo{booktitle}{Proceedings of the 29th ACM International Conference on
  Multimedia}, pp. \bibinfo{pages}{2205--2213}.
\bibitem[{Zhang et~al.(2022)Zhang, Hooi, Hong and Feng}]{zhang2022self}
\bibinfo{author}{Zhang, Y.}, \bibinfo{author}{Hooi, B.}, \bibinfo{author}{Hong,
  L.}, \bibinfo{author}{Feng, J.}, \bibinfo{year}{2022}.
\newblock \bibinfo{title}{Self-supervised aggregation of diverse experts for
  test-agnostic long-tailed recognition}, in: \bibinfo{booktitle}{Advances in
  Neural Information Processing Systems}, pp. \bibinfo{pages}{34077--34090}.

\end{thebibliography}

\end{document}